\documentclass[10pt,twocolumn,letterpaper]{article}

\usepackage{iccv}              

\usepackage{ulem}
\usepackage{multicol}
\usepackage{multirow}
\usepackage{algorithm}
\usepackage{algpseudocode}
\usepackage{colortbl}
\usepackage{graphicx}
\usepackage{booktabs}
\usepackage{stmaryrd}
\usepackage{trimclip}
\usepackage{bm}
\usepackage{amssymb}
\usepackage{pifont}
\usepackage{mathtools}
\newcommand{\shortto}{\rightarrow}
\definecolor{cvprblue}{rgb}{0.21,0.49,0.74}
\definecolor{firstcol}{RGB}{255, 204, 204}  
\definecolor{secondcol}{RGB}{255, 242, 204} 
\definecolor{thirdcol}{RGB}{255, 249, 229}  
\definecolor{iccvblue}{rgb}{0.21,0.49,0.74}
\usepackage[pagebackref,breaklinks,colorlinks,allcolors=iccvblue]{hyperref}

\def\paperID{4771} 
\def\confName{ICCV}
\def\confYear{2025}

\title{DCHM: Depth-Consistent Human Modeling for Multiview Detection}
\author{Jiahao Ma\textsuperscript{1,2},
Tianyu Wang\textsuperscript{2}
Miaomiao Liu\textsuperscript{2}, David Ahmedt-Aristizabal\textsuperscript{2}, Chuong Nguyen\textsuperscript{2} \\
Australian National University\textsuperscript{1}, CSIRO Data61\textsuperscript{2} \\
 {\tt\small \{jiahao.ma, tianyu.wang, miaomiao.liu\}@anu.edu.au}\\ {\tt\small\{david.ahmedtaristizabal, chuong.nguyen\}@data61.csiro.au}
}

\begin{document}
\twocolumn[{%
\renewcommand\twocolumn[1][]{#1}%
\maketitle
\begin{center}
    \centering
    \captionsetup{type=figure}
    \centering
\includegraphics[width=0.9\textwidth]{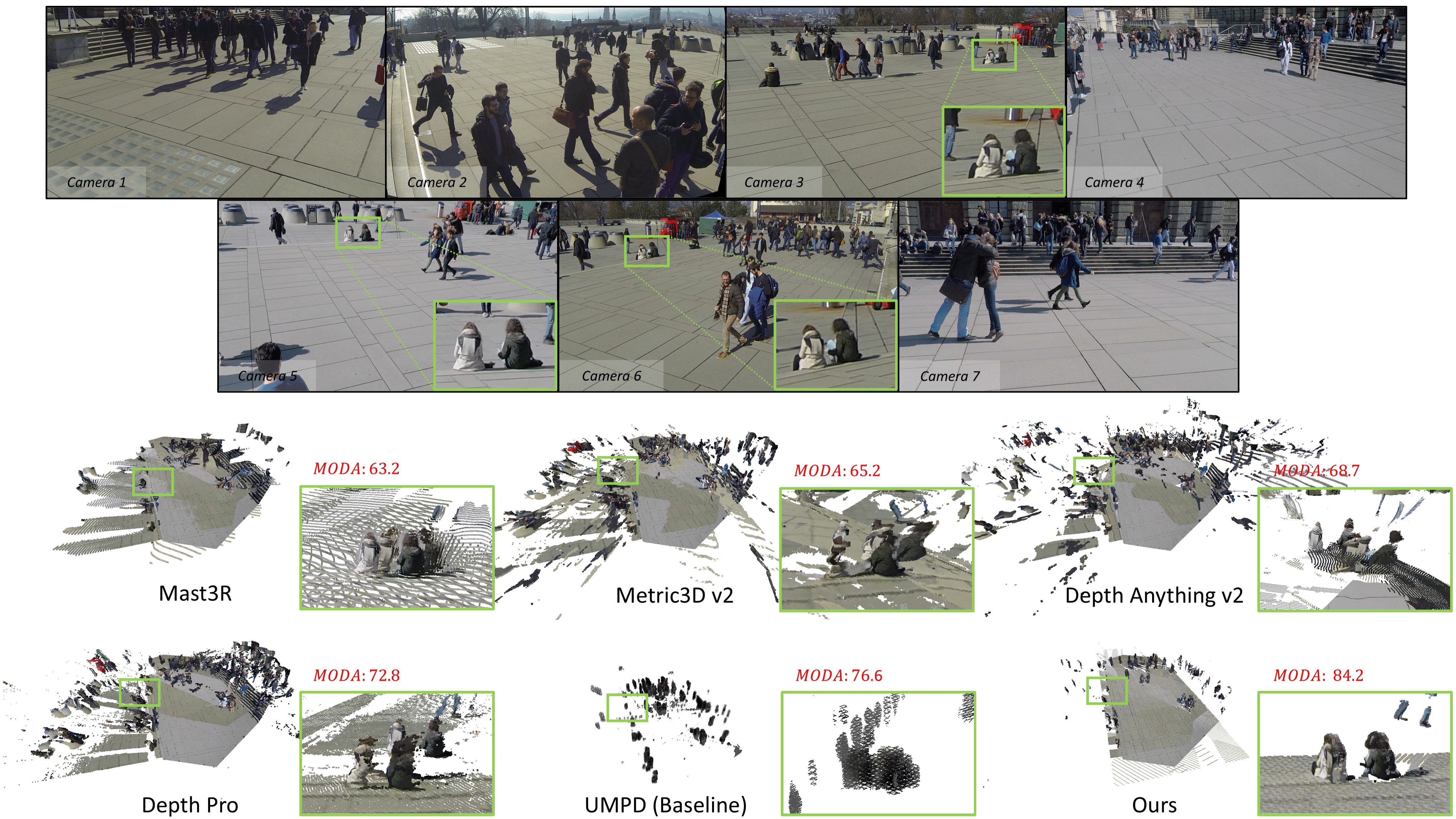}
\caption{
\textbf{Depth-Consistent Human Modeling}. We propose depth-consistent human modeling in multi-view pedestrian detection by leveraging monocular depth estimation to fuse multi-view information into a global coordinate system represented by point clouds. These fused point clouds serve as inputs for pedestrian localization. 
Stereo-based methods~\cite{mast3r} struggle with sparse views and severe occlusions, causing inaccurate geometry predictions. 
Monocular depth estimation methods~\cite{depth_anything_v2, metric3d, depthpro} often lack multi-view consistency, leading to misaligned point clouds—\textit{e.g.}, two people captured by three cameras are projected as separate targets. Baseline UMPD~\cite{umpd} struggles with sparse view reconstruction; erroneous  projections increase false positives. In contrast, our approach ensures spatially consistent depth predictions across views, enabling more accurate detection, as indicated in better MODA scores~\cite{detect_evaluation}. 
}
\label{fig:cover}

\end{center}
}]

\begin{abstract}

Multiview pedestrian detection typically involves two stages: human modeling and pedestrian localization. 
Human modeling represents pedestrians in 3D space by fusing multiview information, making its quality crucial for detection accuracy. However, existing methods often introduce noise and have low precision. While some approaches reduce noise by fitting on costly multiview 3D annotations, they often struggle to generalize across diverse scenes. To eliminate reliance on human-labeled annotations and accurately model humans, we propose Depth-Consistent Human Modeling (DCHM), a framework designed for consistent depth estimation and multiview fusion in global coordinates. Specifically, our proposed pipeline with superpixel-wise Gaussian Splatting  achieves multiview depth consistency in sparse-view, large-scaled, and crowded scenarios, producing precise point clouds for pedestrian localization. Extensive validations demonstrate that our method significantly reduces noise during human modeling, outperforming previous state-of-the-art baselines. Additionally, to our knowledge, DCHM is the first to reconstruct pedestrians and perform multiview segmentation in such a challenging setting. Code is available on the \href{https://jiahao-ma.github.io/DCHM/}{project page}.



\end{abstract}    
\vspace{-10pt}
\section{Introduction}

Multiview pedestrian detection refers to detecting people using images from multiple viewpoints. This setup is particularly beneficial in heavily occluded scenes. However, combining information from multiple views remains challenging due to sparse viewpoints and severe occlusions.

Current methods in this field commonly follow a ``human modeling + pedestrian localization'' strategy.
For human modeling, existing methods project either image features~\cite{mvdet, shot, mvdetr, vfa, 3DROM} or perception results (\textit{e.g.} bounding boxes or segmentation)~\cite{mvchm, xu2016multi, deepocclusion}, onto ground plane to represent pedestrians, followed by neural network regression for pedestrian localization. 
However, these methods suffer from inaccurate feature projections due to the absence of pedestrian height, causing misalignment of points above the ground plane. 
They mitigate projection errors by training from 3D annotation labels. 

To reduce the label dependency and improve generalization across diverse scenes, recent research has explored label-free multiview pedestrian detection without 3D annotations~\cite{Lima_2021_CVPR,zhu2019multi,lopez2022semantic,umpd}. 
They either project heuristic standing points onto the ground via homography projection for multiview fusion~\cite{Lima_2021_CVPR,zhu2019multi,lopez2022semantic} or adopt neural volume rendering~\cite{umpd} for human modeling. While removing the requirement on 3D annotations, they are sensitive to pixel-level errors, where minor pixel-level errors can cause significant projection discrepancies~\cite{Lima_2021_CVPR,zhu2019multi,lopez2022semantic}, or struggles in sparse-view and crowded scenes~\cite{umpd}, often resulting in inconsistent projections, increased false positives, and reduced localization accuracy.

To address this,  we focus on developing label-free methods and propose a more accurate human modeling framework comprising two steps: \textit{reconstruction} and \textit{segmentation}, where reconstruction aligns the 2D segmentation and representation of pedestrians across views. 
 The human modeling represented by segmented Gaussians is subsequently used for more accurate pedestrian localization.

\begin{figure}[t]
  \centering{
    \includegraphics[width=0.9\linewidth]{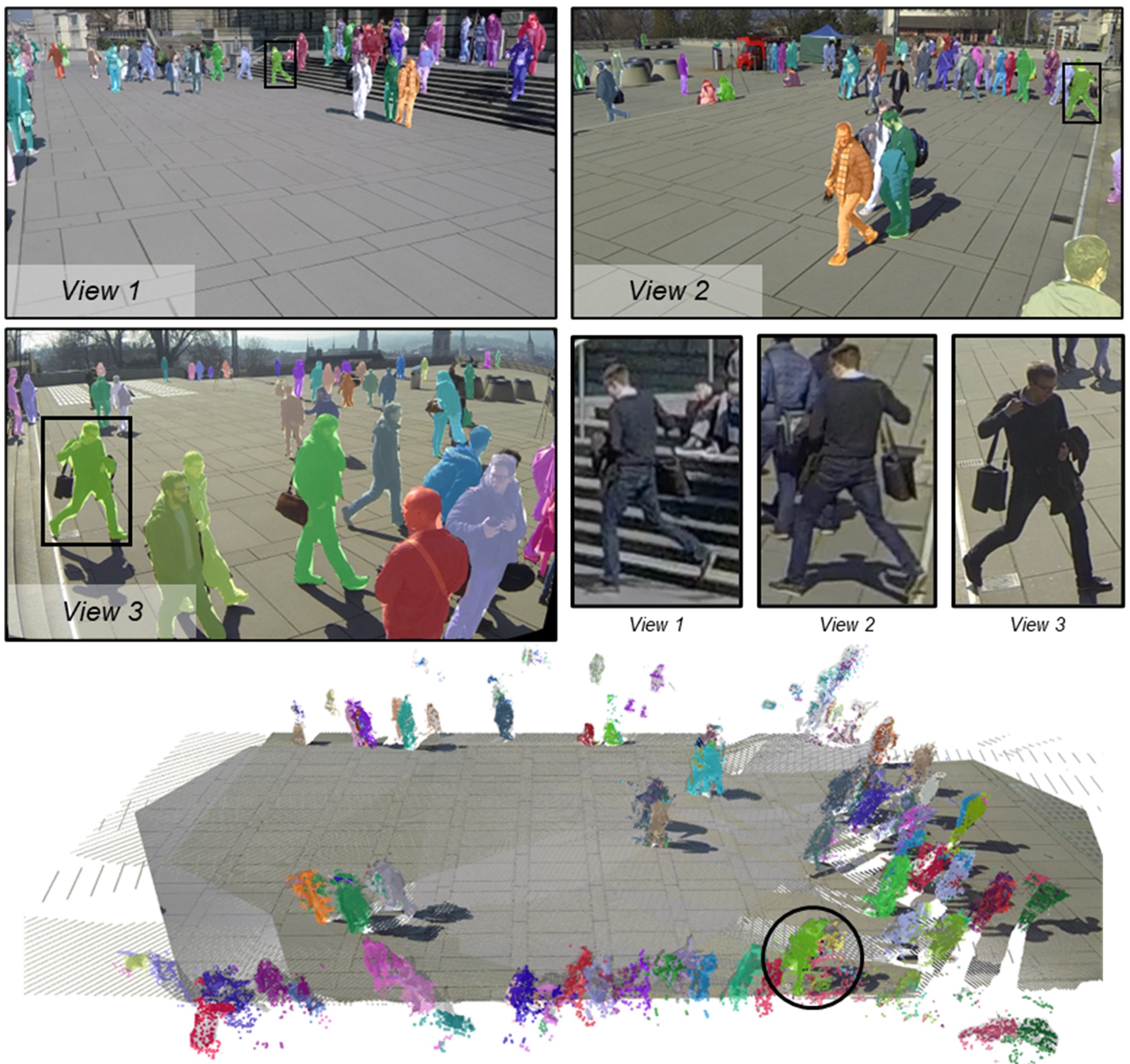}}
  \caption{\textbf{Modeling humans as segmented Gaussians.} Reconstruction and multiview segmentation are performed in \textit{sparse-view}, \textit{large-scale}, and \textit{occluded} scenes.}
  \label{fig:recon_segment}
\vspace{-10pt}
\end{figure}

\begin{figure*}[!t]
  \centering{
    \includegraphics[width=0.95\linewidth]{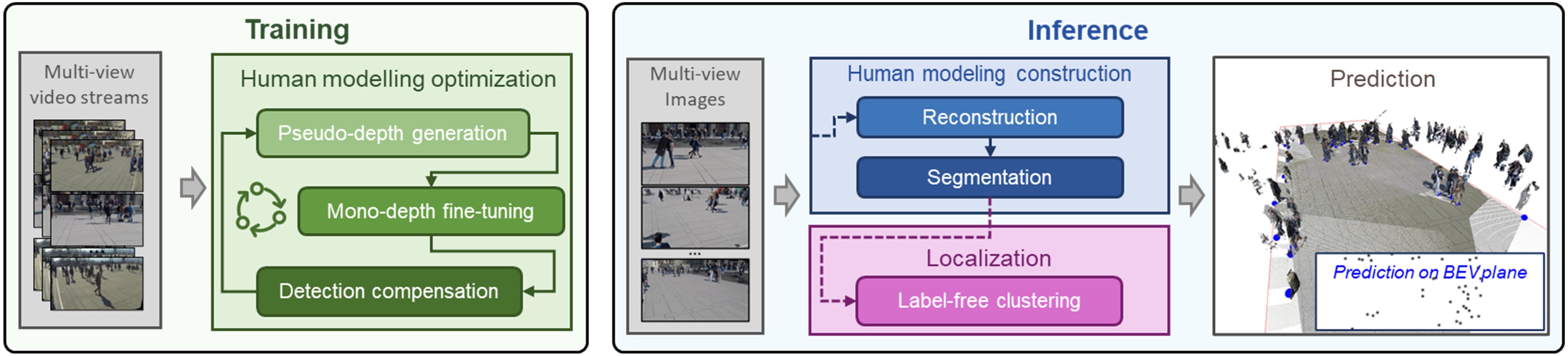}}
  \caption{\textbf{Overview of the framework}. The proposed multiview detection pipeline consists of separate training (left) and inference (right) stages. During training (Sec.~\ref{sec:human_modeling}), human modeling optimization refines mono-depth estimation for multiview consistent depth prediction, via pseudo-depth generation, mono-depth fine-tuning, and detection compensation. During inference (Sec.~\ref{sec:inference}), the optimized mono-depth produces 3D point cloud to model human that are segmented and clustered to detect pedestrians, shown as {\color{blue}{blue points}} on the BEV plane.
  }
  \label{fig:pipeline}
\vspace{-8pt}
\end{figure*}

However, segmentation and reconstruction of crowded scene from images captured at sparse viewpoints are challenging due to severe occlusions and minimal overlap across views, as shown in~Figure.~\ref{fig:cover} and~\ref{fig:recon_segment}. Existing 3D segmentation methods~\cite{gaga, saga, embodiedsam, Gaussian_grouping, flashsplat} rely on densely overlapping views to map 2D segments into 3D space and struggle under sparse-view conditions (see Figure.~\ref{fig:recon_segment}, when only a pedestrian's front and back are visible).
%
Moreover, given sparse multiview images, existing multi-view stereo methods~\cite{colmap} fail to reconstruct the 3D scene as wide camera spacing and sparse viewpoints hinder feature matching. Thus, monocular depth estimation combined with fusion in a global coordinate system for 3D modeling is the most suitable solution.~While existing monocular depth estimation methods~\cite{depth_anything, Marigold, depthpro, metric3d} offer detailed depth from single images, backprojecting each inferred depth map to the 3D space leads to misaligned point cloud for 3D human modeling as those depth maps lack cross-view consistency (see Figure.~\ref{fig:cover}). It then leads to false positives in pedestrian localization.



To address this, we propose to explore depth-consistent 3D human modeling from monocular depth estimation leveraging self-learned multiview consistent pseudo-depth labels.~Specifically, we propose to self-learn depth labels for pedestrian regions using Gaussian Splatting (GS) method~\cite{3dgs}. To overcome the limitation of GS relying on good 3D intialisation, we propose a novel 3D superpixel-based initialization method to allow GS optimization from sparse-view images. Note that GS can only generate reliable depth estimates as pseudo-depth labels for human instances visible in multiview images, which can be achieved via our proposed depth-filtering process relying on geometric consistency check across views. Given the self-learned depth labels, we can fine-tune any monocular depth estimation framework to obtain multiview consistent depth estimation, leading to depth-consistent 3D human modeling. 

Leveraging accurate geometry, we can compensate for 2D missed detections which further allow us to achieve an iterative learning process for improving the geometry with the former depth estimation process. Given the inferred geometry, we introduce multiview label matching, ensuring that the same pedestrian is consistently mapped across different views for 3D segmentation. Finally, we achieve pedestrian localization by clustering the segmented 3D Gaussians. 
Our well-designed pipeline achieves state-of-the-art results compared to existing label-free methods.
%
%

\noindent In summary, our main contributions are as follows: 
\begin{itemize}
    \item[i)] \textit{DCHM} employs Gaussian Splatting with superpixels-wise optimisation to iteratively refine monocular depth estimation using geometric constraints, enabling accurate scene reconstruction for enhanced human modeling.
    %
    %
    \item[ii)] \textit{DCHM} introduce multiview label matching to segment reconstructed geometry, ensuring consistent 2D/3D pedestrian segmentation in sparse-view, crowded scenes.
    %
    %
    \item[iii)] We propose an integrated detection compensation mechanism that enhances the performance of existing monocular detection methods.
    \item[iv)]The proposed method, evaluated on multiview pedestrian datasets without 3D annotation training, significantly outperforms existing label-free approaches.
\end{itemize}

\section{Related work}
\label{sec:related_work}
\noindent\textbf{Label-based multiview detection}. 
Supervised approaches typically project either image features~\cite{mvdet, shot, mvdetr, 3DROM, earlybird, engilberge2023multi, lee2023multi, cardoen2024label, BGDM, brorsson2024mvuda} or perception outputs (\textit{e.g.}, bounding boxes, segmentation)~\cite{PR, Lima_2021_CVPR, xu2016multi, pomcnn, deepocclusion, mvchm} onto a ground plane or 3D volume~\cite{vfa} for pedestrian localization. Hou \textit{et al}.~\cite{mvdet} introduced a perspective transformation scheme and a fully convolutional network for feature aggregation, while Song \textit{et al}.~\cite{shot} employed stacked homography transformations to capture features at multiple height levels. To mitigate shadow-like distortions, Hou \textit{et al}.~\cite{mvdetr} adopted a transformer architecture, and Ma \textit{et al}.~\cite{vfa} voxelized 3D features for vertical alignment. Other methods leverage occupancy maps\cite{pomcnn}, mean-field inference~\cite{deepocclusion}, and CRFs~\cite{CRF} to fuse detections from multiple views. Building on 2D detections, some approaches rely on logic functions for pedestrian presence and clustering~\cite{PR}, or represent pedestrians as cardboard-like point clouds to regress 3D positions~\cite{mvchm}. Moreover, object skeletons~\cite{selfpose3d, 3dmuppet} have been explored for multiview fusion. Although these strategies reduce projection errors by fitting multiview annotations, they often demand extensive labeling and struggle to generalize to diverse scenes.

\noindent\textbf{Label-free Multiview Detection.} To reduce reliance on labeled data and improve generalization, recent studies~\cite{Lima_2021_CVPR,zhu2019multi,lopez2022semantic,umpd} have explored label-free multiview pedestrian detection without 3D annotations. Lima \textit{et al}.~\cite{Lima_2021_CVPR} estimate pedestrian standing points in 2D bounding boxes and address a clique cover problem to infer 3D coordinates. UMPD~\cite{umpd} maps 2D to 3D by predicting pedestrian densities with a volume-based detector, trained using pseudo labels from SIS. Yet, these approaches struggle in sparse-view and crowded scenes, leading to inconsistent projections and reduced localization accuracy.

\noindent\textbf{Gaussian-based 3D segmentation}.
Segment Anything (SAM)~\cite{sam} has laid the foundation for open-world segmentation. Integrating Gaussian Splatting with SAM enables flexible 3D segmentation and scene editing. SA-GS~\cite{saga} renders 2D SAM feature maps with a SAM-guided loss for 3D segmentation, while Feature 3DGS~\cite{feature3dgs} distills LSeg~\cite{LSeg} and SAM features into 3D Gaussians. However, these methods often suffer from inconsistent segmentation due to incoherent masks across views. 
Gaussian Group~\cite{Gaussian_grouping} and CoSSegGaussians~\cite{CoSSegGaussians} attempt to address this by using 2D segmentation masks as pseudo-labels with object tracking and linking masks across views. Yet, they struggle with sparse view correspondence. In this paper, we tackle highly occluded, sparse-view scenes lacking coherent masks using Gaussians for robust 3D clustering, resulting in more consistent segmentation in complex environments.

\subsection{Preliminary: 3D Gaussian Splatting}
\label{preliminaries}
Gaussian Splatting (GS)~\cite{3dgs} represents 3D scenes using a set of 3D Gaussians, each defined by $(\bm{\mu}, \bm{\Sigma}, o, \mathbf{sh})$, where $\bm{\mu} \in \mathbb{R}^3$ is the spatial center, $\bm{\Sigma}$ the covariance matrix, $o \in [0, 1]$ the opacity, and $\mathbf{sh}$ the view-dependent color. Rendering involves projecting Gaussians onto the image plane as 2D Gaussians:
\begin{equation}
G'(\mathbf{u}) = e^{ -\frac{1}{2} (\mathbf{u} - \bm{\mu}')^\intercal \bm{\Sigma}'^{-1} (\mathbf{u} - \bm{\mu}') },
\end{equation}
where $\mathbf{u} \in \mathbb{R}^2$ is the pixel coordinate, $\bm{\mu}'$ the projected center, and $\bm{\Sigma}'$ the 2D covariance matrix. The Gaussians, sorted by depth, are alpha-blended to compute per-pixel features:
\begin{equation}
\resizebox{.9\linewidth}{!}{
$\displaystyle
F(\mathbf{u}) = \sum_{i=1}^N T_i \alpha_i f_i, \quad
\alpha_i = o_i G'_i(\mathbf{u}), \quad
T_i = \prod_{j=1}^{i-1} \bigl(1 - \alpha_j\bigr).
$}
\end{equation}

The rendering outcome depends on $f$: view-dependent color produces an RGB image, Gaussian center depth yields a depth frame, and $f = 1$ produces a soft binary mask.
\vspace{-7pt}
\begin{figure*}[t]
  \centering{
    \includegraphics[width=0.95\linewidth]{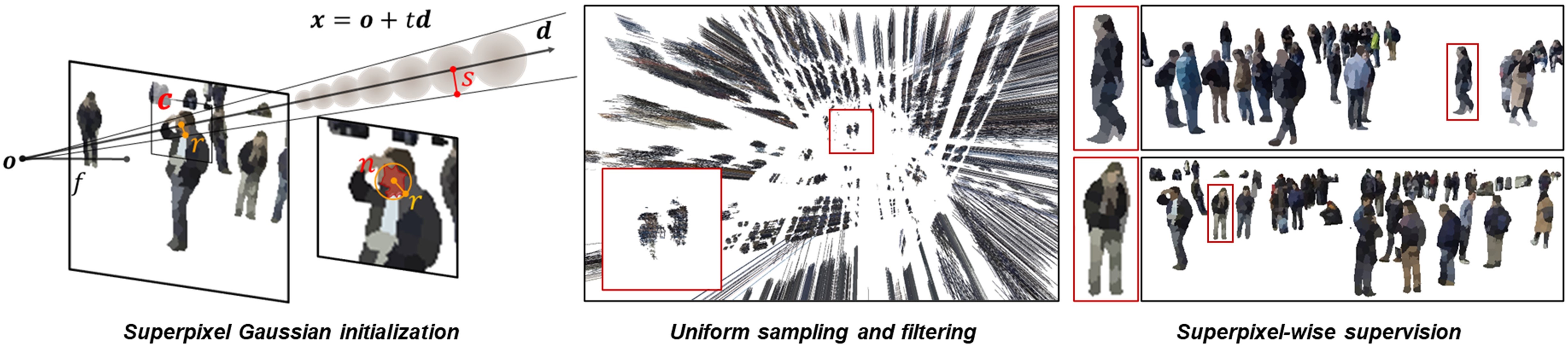}}
\vspace{-5pt}
  \caption{Superpixel-wise Gaussian initialization and optimization. 
  }
  \label{fig:gs_construction}
\vspace{-12pt}
\end{figure*}

\section{Method} 
Our goal is to achieve depth-consistent 3D human modeling (DCHM) by estimating multiview consistent depth maps $\{D^v\}_{v\in \mathcal{V}}$ from sparse view images
$\{I^v\}_{v\in \mathcal{V}}$ captured by calibrated cameras and 2D pedestrian segmentation masks $\{M^v\}_{v\in \mathcal{V}}$, where $\mathcal{V}$ denotes the set of sparse views.~The estimated point cloud will be converted to segmented Gaussians for localization.~The overview of our method is shown in Figure.~\ref{fig:pipeline}. Our proposed framework involves iterative training of three stages, pseudo depth labels estimation from GS, monocular depth model fine-tuning and multi-view detection compensation. In the following, we begin by reviewing the preliminaries of GS, followed by a detailed description of the reconstruction and segmentation modules for DCHM.




\subsection{Depth-Consistent 3D Human Modeling}
\label{sec:human_modeling} 
Our proposed depth-consistent 3D human modeling is achieved by reconstruction from multiview consistent depth maps $\{D^v\}_{v\in \mathcal{V}}$ and 3D segmentation given the modeled geometry. To this end, we propose an iterative training method to self-learn pseudo depth labels from the sparse-view images using GS which are used to fine-tune any monocular depth estimation method to obtain $\{D^v\}_{v\in \mathcal{V}}$.

\subsubsection{Consistent pseudo-depth label generation using GS}
\label{sec:pseudo_depth_generation}
Optimizing GS~\cite{zhu2024fsgs, xiong2024sparsegs, xu2024mvpgs} in a sparse-view setting is nontrivial. GS relies on pixel-wise photometric loss for geometry optimization, but limited viewpoint overlap complicates this process. Additionally, pixel color variation in overlapping regions--caused by differences in lighting and camera placement--lead to inconsistent reconstructions and potential failure. 
To address this, we represent each pedestrian with $K$ superpixel (Figure~\ref{fig:gs_construction}.~\textit{right}), which aggregate local features to enhance consistency in overlapping regions. Superpixel-wise supervision not only improves alignment but also accelerates optimization, producing a coarse yet reliable geometry for accurate localization. Given a synchronized multicamera setup with known poses, we reconstruct the foreground and ground separately (the latter using camera parameters and a predefined depth range, see supplementary). The main challenge, however, lies in accurately reconstructing the foreground, as detailed below.


\noindent \textbf{Superpixel-wise Gaussian initialization}.
Optimising GS normally relies on good 3D points as initialisation. However, due to wide camera baselines in our setup, traditional Structure-from-Motion (SfM)~\cite{colmap} fails to reconstruct sparse point clouds for initialization. We address this by a ``uniform sampling + filtering'' strategy.  We cast rays through the center of each superpixel and uniformly sample points along each ray, following  
$\mathbf{x} = \mathbf{o} + t \mathbf{d}$,
where $\mathbf{x}$ denotes the position, $\mathbf{o}$ is the ray origin, and $\mathbf{d}$ is the ray direction.  
Each sampled point is initialized as a Gaussian, with its position as the mean $\bm{\mu}$, the superpixel color as the base color of $\mathbf{sh}$, and an initial opacity of 0.01. The Gaussian is initialized as a sphere with a uniform orientation and a scale related to the superpixel area and distance $t$ between $\bm{\mu}$ and $\mathbf{o}$. The superpixel area is approximated by a circle on the image plane. As shown in Figure~\ref{fig:gs_construction}.~\textit{left}, the radius of the circle can be calculated by  $r=\sqrt{n \cdot \Delta x \cdot \Delta y / \pi}$, where $n$ denotes the number of pixels in the superpixel, and $\Delta x$ and $\Delta y$ represent the pixel width and height in world coordinates, obtained from the calibrated camera parameters. The scale $s$ of Gaussian can be computed by
\vspace{-5pt}
\begin{equation}
\resizebox{.75\linewidth}{!}{
$\displaystyle
    s = \frac{t f r}{\|\mathbf{c} - \mathbf{o}\|_2 \cdot \sqrt{\left(\sqrt{\|\mathbf{c} - \mathbf{o}\|_2^2-f^2}- r \right)^2+f^2}.} 
$}
\label{eq:radius}
\end{equation}
\noindent
Here, $f$ is the focal length and $\mathbf{c}$ is superpixel center in world coordinates. After initialization, Gaussians are then projected onto views, discarding those in background masks (Figure~\ref{fig:gs_construction}.~\textit{middle}).


\noindent \textbf{GS optimization}. 
We optimize, prune, and grow the Gaussians based on the parameter gradients with respect to our loss $\mathcal{L}$. 
$\mathcal{L}$ is defined as a weighted sum of four components: superpixel-wise photometric loss $\mathcal{L}_{sp}$, mask loss $\mathcal{L}_m$, opacity loss $\mathcal{L}_o$, and depth constraint loss $\mathcal{L}_d$:
\begin{equation}
\mathcal{L} = \lambda_{sp} \mathcal{L}_{sp} + \lambda_m \mathcal{L}_m + \lambda_d \mathcal{L}_d + \lambda_o \mathcal{L}_o,
\end{equation}
with $\lambda_{sp}$, $\lambda_m$, $\lambda_d$, and $\lambda_o$ being scaler weights.
To compute the loss, for each view $v$ we render from GS the foreground mask $\hat{M}^v$, the superpixel image $\hat{I}^v$ and depth $D^v$ as introduced in Sec.~\ref{preliminaries}. We further denote the superpixel image as $I^v$ generated from ground truth RGB image, as shown in Figure~\ref{fig:gs_construction}.~\textit{right}. For pedestrian masks generated by segmentation networks, $M_i^v$ is the binary mask of the $i^{th}$ pedestrian and $M^v = \sum_{i=1} M^v_i$. The sum is element-wise.
We use $\mathbf{u}$ to denote individual pixels. 
$F(\mathbf{u})$ is the value of feature map $F$ at pixel $\mathbf{u}$ for $F \in \{M_i^v, M^v, \hat{M}^v, I^v, \hat{I}^v, D^v\}$.

\noindent \textit{Superpixel-wise photometric loss} $\mathcal{L}_p$ measures the $L_1$ distance between the rendered image and the ground truth and is defined as:
\begin{equation}
   \mathcal{L}_{sp} = \sum_{v} \sum_{\mathbf{u}: M^v(\mathbf{u}) = 1} \| I^v(\mathbf{u}) - \hat{I}^v(\mathbf{u}) \|_1.
\end{equation}

\noindent \textit{Mask loss} $\mathcal{L}_m$ computes the difference between the $M^{v}$ and rendered mask $\hat{M}^v$ as:
\begin{equation}
   \mathcal{L}_m = \sum_{v} \sum_{\mathbf{u}}\| M^{v}(\mathbf{u}) - \hat{M}^v(\mathbf{u}) \|_2^2.
\end{equation}
\noindent  \textit{Depth constraint loss} $\mathcal{L}_d$.We encourage a small variance of depth predictions for pixels inside each pedestrian mask:
\begin{equation}
\resizebox{.65\linewidth}{!}{
$\displaystyle
\mathcal{L}_{d} = \sum_{v} \sum_{i} \sum_{\mathbf{u}: M_i^v(\mathbf{u}) = 1} \! \! \! \! \! \! \! \! \| D^v({\bf u}) - \bar{d}^v_i \|^2_2,
$}
\label{eq:loss_d}
\end{equation}
\begin{equation}
\resizebox{.89\linewidth}{!}{
$\displaystyle
\bar{d}^v_i := (|M_i^v|)^{-1} \smashoperator{\sum_{\mathbf{u}: M_{i}^v(\mathbf{u}) = 1}} D^v({\bf u}),\ \ \  |M^v_i| := \sum_{\mathbf{u}} M^v_i(\mathbf{u}).
$}
\end{equation}
$\mathcal{L}_d$ prevents the large depth disparities within a single mask.

\noindent \textit{Opacity loss} encourages the opacity $o_i$ of each Gaussian to be close to zero or one, promoting a clear distinction between foreground and background~\cite{gaussian_surfels}, defined below: 
\begin{equation}
\mathcal{L}_o = \sum_i e^{-(o_i - 0.5)^2 / 0.04}.
\end{equation}

\begin{figure}[t]
  \centering{
    \includegraphics[width=0.99\linewidth]{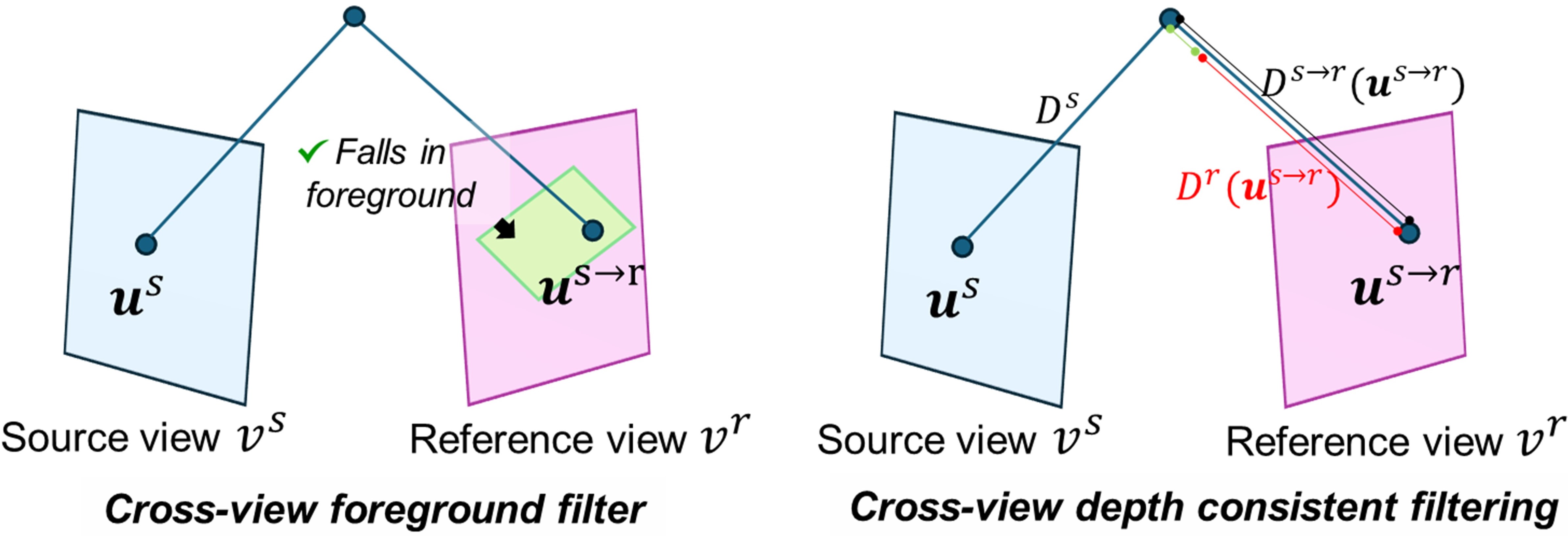}}
  \caption{Pseudo-depth filtering strategies. 
  }
  \label{fig:filter}
\vspace{-12pt}
\end{figure}

\noindent\textbf{Pseudo depth filtering}.
GS could obtain reliable depth labels only for humans visible in multiviews. Thus, after optimization, we propose pseudo-depth filtering strategies to retain valid depth labels for the following monocular depth model fine-tuning. Depth values are validated based on two conditions: reprojected depth must 1) remain in the foreground region \textit{across all views}, and 2) be consistent with GS-rendered depth \textit{in at least one other view}.

To enforce this, we introduce \textit{Cross-view foreground filtering} and \textit{Cross-view depth consistency filtering} as post-processing steps, detailed in Figure~\ref{fig:filter}. Given a source view $v^s$ and a reference view $v^r$, let $\mathbf{u}^{s}$ be the homogeneous coordinate of a pixel in $v^s$, with depth $D^{s}(\mathbf{u})$ in the depth frame $D^{s}$.
We define $\mathbf{u}^{s \shortto r}$ as the reprojected homogeneous coordinate of $\mathbf{u}^{s}$ in $v^r$, and $D^{s \shortto r}$ as the reprojected depth, calculating using intrinsic and extrinsic camera parameters (See supplementary for details).
Each view serves as the source view while all others act as references during the filtering processes:
\begin{itemize}
    \item \textit{Cross-view foreground filtering}.
    In $v^s$, for each foreground pixel $\mathbf{u}^s$, we compute $\mathbf{u}^{s \shortto r}$ and discard $\mathbf{u}^s$ if $\mathbf{u}^{s \shortto r}$ falls into the background in any reference view.
    \item \textit{Cross-view depth consistency filtering}. 
    For each reference view $v^r$ we compute $\mathbf{u}^{s \shortto r}$ and $D^{s \shortto r}$. Then we compare the projected depth $D^{s \shortto r}(\mathbf{u}^{s \shortto r})$ against the rendered depth  $D^r(\mathbf{u}^{s \shortto r})$ for each pixel $\mathbf{u}^s$ in the source view. We retain $\mathbf{u}^s$ if $|D^{s \shortto r}(\mathbf{u}^{s \shortto r}) - D^r(\mathbf{u}^{s \shortto r})| < \tau$ for some $v^r$. Here $\tau$ is a pre-defined threshold.
\end{itemize}

\begin{figure*}[t]
  \centering{
    \includegraphics[width=0.99\linewidth]{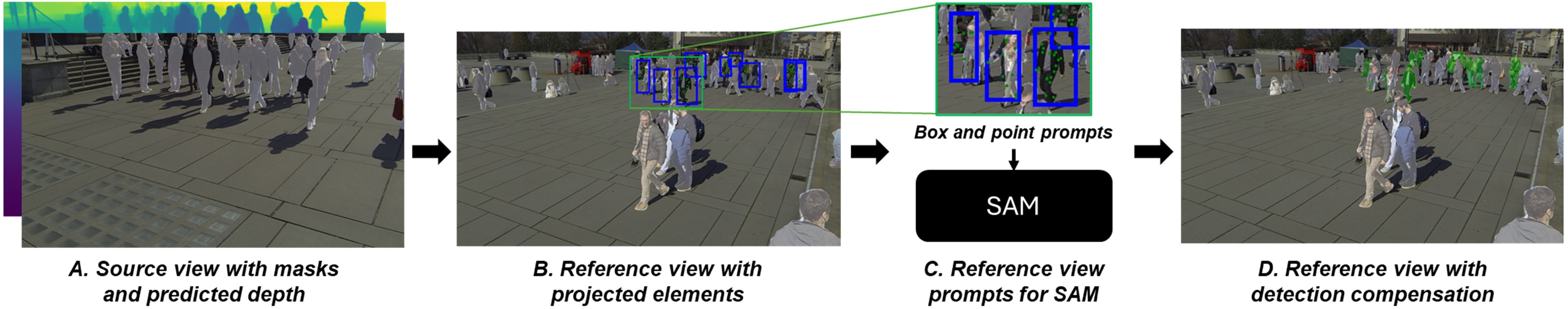}}
  \caption{\textbf{Multi-view compensation for missed detection}. (A) Source-view masks to be projected into the reference view by predicted depth; (B) the projected elements form point-box prompts for SAM; (C) these prompts facilitate segmentation; (D) yielding compensatory detections in the reference frame.}
  \label{fig:label_compensation}
\vspace{-10pt}
\end{figure*}

The depth of retained pixels after filtering then serves as pseudo-depth labels for subsequent mono-depth estimation. 

\subsubsection{Monocular depth fine-tuning} 
\label{sec:depth_fine_tuning}
Our pseudo-depth labels can be leveraged to fine-tune any off-the-shelf monocular depth estimation network. Fine-tuning with pseudo-depth supervision progressively improves the network's valid depth predictions, as shown in rows 2 and 3 of Figure~\ref{fig:pseudo_label}.


\subsubsection{Multi-view detection compensation} \label{sec:compensation}
Occlusions can cause missed detections, which may impact the performance of our method. To mitigate this, we introduce a compensation mechanism during optimization to recover missed pedestrian detections. 
As shown in Figure~\ref{fig:label_compensation}, we select one camera view as the \textit{reference view} and the remaining views as \textit{source views}. Using the predicted masks and depth information from the source views, we project the foreground (pedestrians) onto the reference view. Box prompts computed from the projection bounds and point prompts initialize mask by SAM~\cite{sam}, followed by prompt refinement through mask-guided sampling. The refined prompts, represented by the blue boxes and green points in Figure~\ref{fig:label_compensation}.~C, are then used to obtain the final segmentation mask via SAM.
The process is repeated for each camera view to ensure comprehensive compensation across all viewpoints. Refer to supplementary material for details.


\subsubsection{Iterative improvement}
\label{iterative_improvement}
\begin{figure}[t]
  \centering{\includegraphics[width=0.9\linewidth]{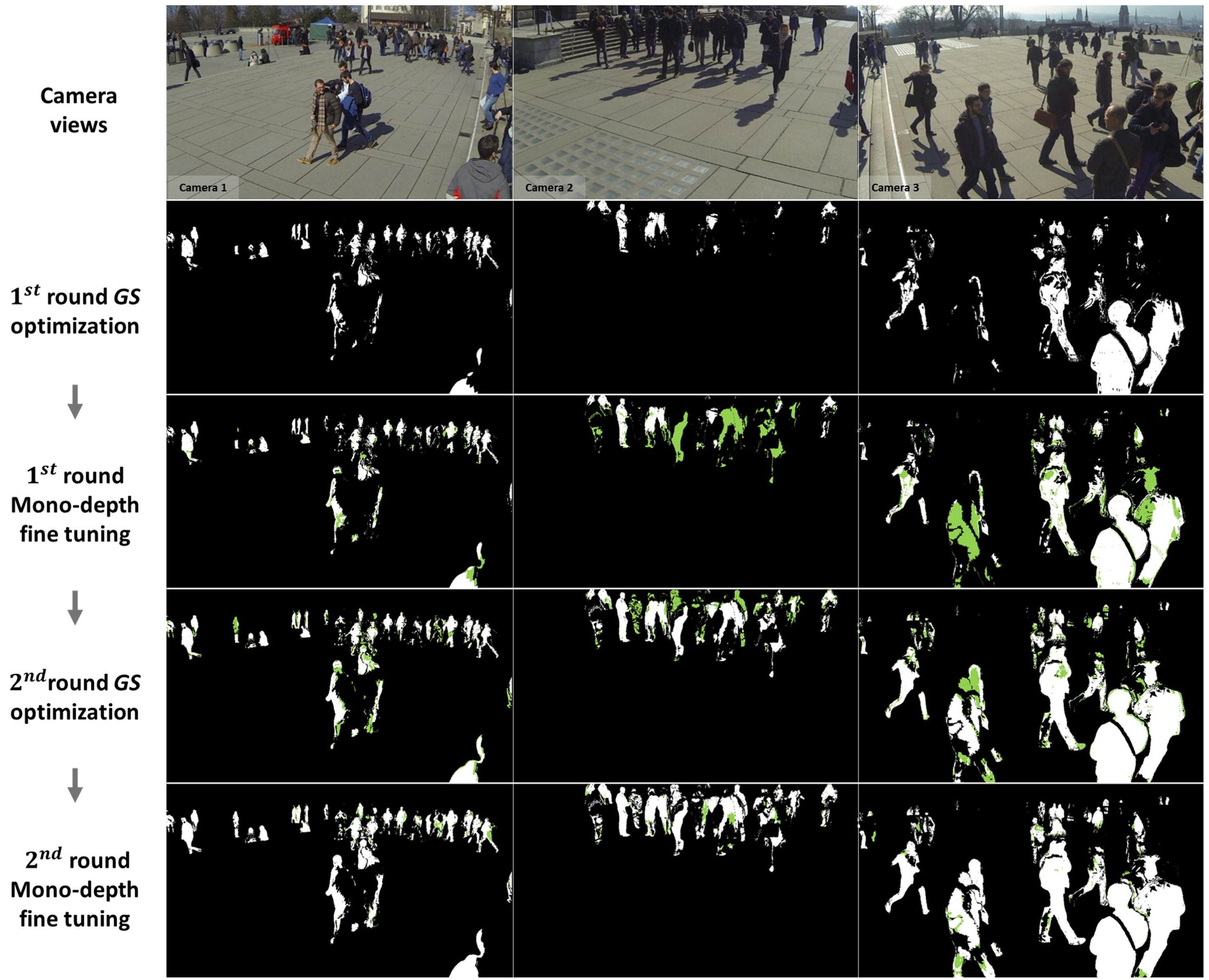}}
  \caption{\textbf{Increasing valid pseudo-depth during iterative learning.} Binary masks illustrate the valid pseudo-depth regions, with white color indicating valid areas and black color denoting invalid ones. Green highlights newly added valid depth regions compared to the previous optimization cycle.}
  \label{fig:pseudo_label}
\vspace{-17pt}
\end{figure}

The pseudo-depth generation using GS, fine-tuning of mono-depth estimation, and multiview detection compensation create an iterative training loop. Gaussian optimization generates pseudo-labels to supervise monocular depth estimation, and the fine-tuned network's updated depth predictions serve as initialization for the next Gaussian optimization cycle. Furthermore, the refined depth compensates for the missed detection for better GS optimization. As illustrated in Figure~\ref{fig:pseudo_label}, the expanding green regions highlight this iterative refinement, where each cycle adds valid depth estimates that enhance subsequent rounds.

\subsection{Inference}
\label{sec:inference}
In inference, the proposed pipeline consists of two main steps: human modeling construction and localization.

\label{sec:unsup_loc}

\begin{figure}[t]
  \centering{
    \includegraphics[width=0.85\linewidth]{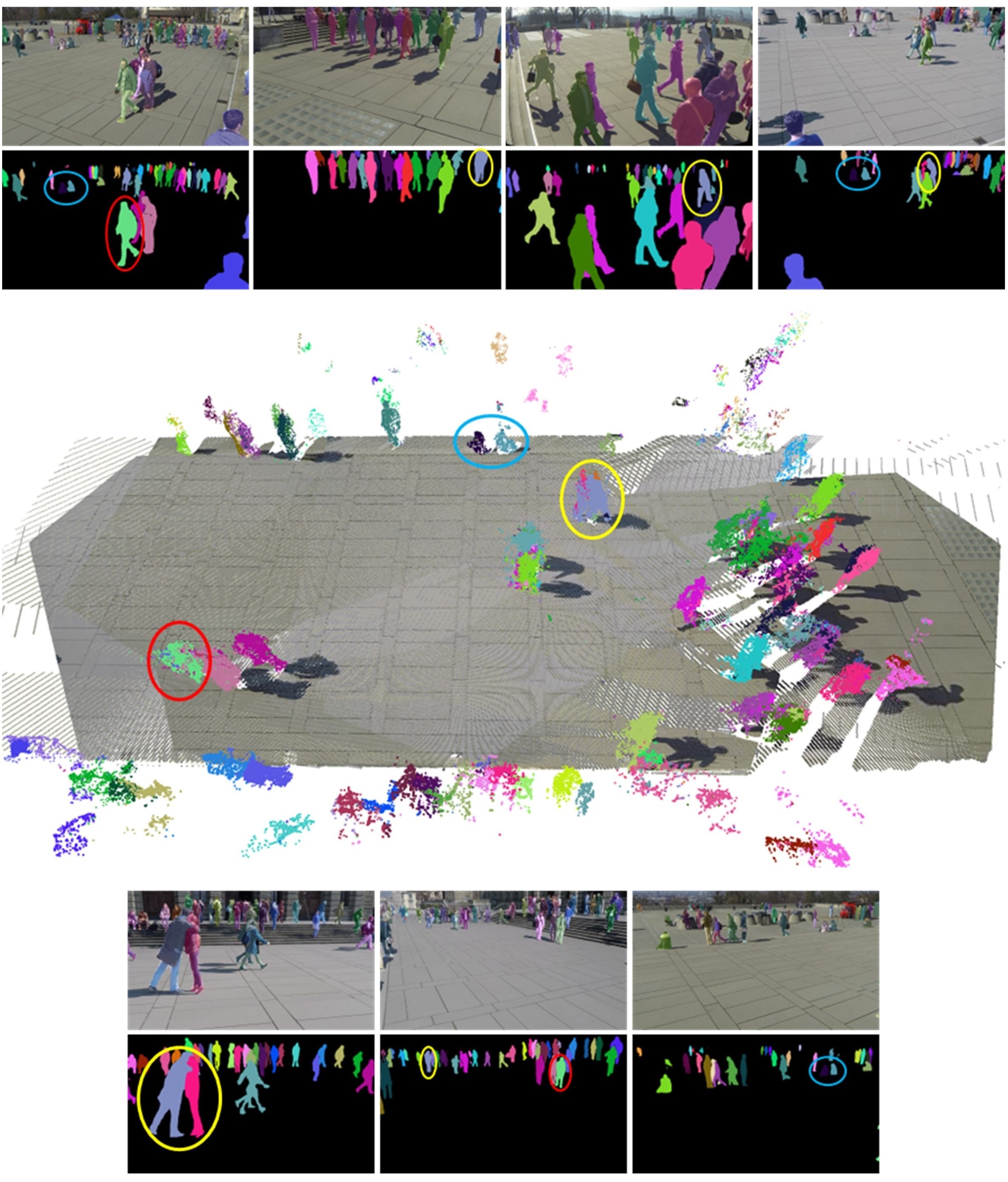}}
  \caption{\textbf{Multi-view pedestrian label matching and label-free clustering}. Pedestrians, represented by segmented Gaussians, are assigned unique IDs indicated by distinct colors. Matching occurs in 3D and is projected onto 2D image marks across camera views, with correspondences highlighted in colored circles. }
  \label{fig:clustering}
\vspace{-15pt}
\end{figure}

\subsubsection{Human modeling construction}
\noindent\textbf{Reconstruction}.
We use an optimized monocular depth estimator to project depth onto a global point cloud converted to Gaussians--where the pixel RGB defines color, point position set the means, and camera parameters determine rotation and scale. The human modeling process takes only 0.1 seconds on a GeForce RTX 4090.

\noindent\textbf{Segmentation: multiview pedestrian matching}. 
Single-view segmentation is independent, causing the same person to be mismatched across views. Hence, matching corresponding pedestrian instances across different views is essential for accurate 3D localization.
To address this, we propose multiview pedestrian matching to map cross-views pedestrians by Gaussians (see Figure.~\ref{fig:clustering}). 
First, we project the Gaussians onto a reference view. Given the pedestrian mask $M_i^c$, we assign the same ID to Gaussians that fall within the same mask. The blending weight $w_i = \sum_{i} T_i \alpha_i$ is computed for each Gaussian based on the pixels $u_i$ it covers. If the blending weight $w_i$ exceeds a certain threshold $\tau_{vis}$, we assign the ID to both the mask and the Gaussians within it. This method accounts for occlusion by selecting only those Gaussians located on the visible surface. 

After the initial assignment in the first camera, some Gaussians are assigned an ID, while others do not. We then project the Gaussians onto a second camera and encounter two situations:
(1) if Gaussians with IDs fall within a human mask, we assign the most frequent ID to the current human mask; 
(2) if Gaussians within the human mask lack IDs, we assign a new ID to both the mask and the Gaussians within it. 
This process is repeated for all camera views, ensuring that all Gaussians are assigned IDs and pedestrians are correctly matched across views.
\label{sec:mmatching} 

\subsubsection{Localization}
\noindent\textbf{Label-free clustering}. 
For each human ID and its corresponding Gaussians, we check if the number of Gaussians exceeds a specified threshold $\tau_{cluster}$. If it does, we apply the DBSCAN algorithm~\cite{DBSCAN} to cluster the selected Gaussians and compute the cluster centers. The count of Gaussians within each cluster is used as a measure of confidence. We then apply non-maximum suppression (NMS) to remove redundant clusters. The remaining clusters and their respective centers represent the final predicted pedestrian locations. A detailed summary of the localization method can be found in the supplementary material. As shown in Figure~\ref{fig:clustering}, even under sparse-view and occluded conditions, our approach consistently matches pedestrians across views, along with 3D segmentation and localization.

\section{Experiments}

\subsection{Experimental settings}
We evaluate our method on three benchmarks: MultiviewX~\cite{mvdet}, Wildtrack~\cite{wildtrack}, and Terrace~\cite{terrace}.

\noindent\textbf{MultiviewX}~\cite{mvdet} is a synthetic dataset from Unity simulating crowded scenes over a $16 \times 23$ $\mathrm{m}^2$ area with 6 calibrated cameras at $1080 \times 1920$ resolution. Localization is evaluated on the last 40 frames.

\noindent\textbf{Wildtrack}~\cite{wildtrack} is a real-world benchmark over $12 \times 36$ m$^2$ with 7 cameras recording 35-minute, 60 fps videos at $1080 \times 1920$. We sample 4200 frames per camera for fine-tuning and use 40 labeled frames for evaluation.

\noindent\textbf{Terrace}~\cite{terrace} is a real-world dataset with 4 cameras covering $5.3 \times 5$ m$^2$, with 300 training and 200 validation frames.





\noindent \textbf{Evaluation metrics}. 
We evaluate performance using four metrics: Multiple Object Detection Accuracy (MODA), Multiple Object Detection Precision (MODP)~\cite{detect_evaluation}, Precision, and Recall. MODA measures normalized missed detections and false positives, while MODP assesses localization precision. Precision is computed as $P = \frac{TP}{TP + FP}$ and Recall as $R = \frac{TP}{TP + FN}$. A 0.5-meter distance threshold is applied to determine true positives.

\subsection{Implementation details}
\noindent\textit{Data processing and fine-tuning}. 
In the MultiviewX and Terrace dataset, we used training images for mono-depth fine-tuning. In the Wildtrack dataset, we extract 4200 images per camera from 35-minute videos sampled at 2 frames per second. Each pedestrian has $K = 30$ superpixels, obtained by~\cite{slic}. We use YOLOv11~\cite{yolov11} for segmentation and employ Depth Anything v2~\cite{depth_anything_v2} for depth fine-tuning.

\noindent\textit{Localization}. 
For label-free clustering, we discard Gaussians with a blending weight lower than $0.05$.



\begin{table}[t!]
\begin{center}
    \scriptsize
    \resizebox{0.9\linewidth}{!}{
    \begin{tabular}{l c c c c}
    \toprule
    \multicolumn{2}{c}{Method} & & \multicolumn{2}{c}{\textbf{Wildtrack}} \\
    \cmidrule[0.3pt]{1-2} \cmidrule[0.3pt]{4-5} 
    \textit{Human modeling} & \textit{Localization} & & MODA & MODP \\
    \midrule[0.4pt]
    Marigold~\cite{Marigold}          & \multirow{6}{*}{Our \textit{Loc.}}&  & $58.1$        & $47.0$      \\
    Metric3D v2~\cite{metric3d}       &                                  &  & $65.2$        & $59.8$      \\
    Mast3R~\cite{mast3r}              &                                  &  & $63.2$        & $60.2$      \\
    Depth Anything v2~\cite{depth_anything_v2} &                          &  & $68.7$        & $55.1$      \\
    Depth Pro~\cite{depthpro}          &                                  &  & $\cellcolor{secondcol}{72.8}$        & $\cellcolor{secondcol}{64.1}$      \\
    Our \textit{Recon.}                  &                                  &  & $\cellcolor{firstcol}{84.2}$        & $\cellcolor{firstcol}{80.3}$     \\ 
    \bottomrule
    \end{tabular}
    }
  \end{center}
\vspace{-10pt}
  \caption{Comparison of human modeling via different mono-depth estimation for label-free multiview detection. Human modeling optimized by proposed pipeline (``Our \textit{Recon.}'') outperforms existing depth estimation methods under the same label-free localization (``Our \textit{Loc.}''), highlighting our accurate reconstruction.}
  \label{table:depth_comparison}
\vspace{-10pt}
\end{table}

\begin{table*}[t!]
  \begin{center}
      \scriptsize
      \resizebox{0.95\linewidth}{!}{
      \begin{tabular}{l cccc c cccc c cccc}
      \toprule
      \multirow{2}{*}{Method} & \multicolumn{4}{c}{\bf Wildtrack} & \quad & \multicolumn{4}{c}{\bf Terrace} & \quad & \multicolumn{4}{c}{\bf MultiviewX} \\
      \cmidrule[0.3pt]{2-5} \cmidrule[0.3pt]{7-10} \cmidrule[0.3pt]{12-15}
      & MODA & MODP & Precision & Recall & \quad & MODA & MODP & Precision & Recall & \quad & MODA & MODP & Precision & Recall\\
      \midrule[0.4pt]
      RCNN  \&  clustering~\cite{xu2016multi} &  $11.3$ &  $18.4$  &  $68.0$  &  $43.0$  & \quad & $-11$ & $28$ & $39$ & $50$ & \quad & $18.7$ & $46.4$ & $63.5$ & $43.9$ \\
      POM-CNN~\cite{lopez2022semantic}   &  $23.2$   &  $30.5$   &  $75.0$  & $55.0$  & \quad & $58$ & $46$ & $80$ & $78$ & \quad & - & - & - & -  \\
      Pre-DeepMCD \cite{zhu2019multi} & $33.4$ & $52.8$ & $\cellcolor{firstcol}{93.0}$ & $36.0$ & \quad & - & - & - & - & \quad & - & - & - & - \\
      BP \& BB + CC~\cite{Lima_2021_CVPR}  &  $56.9$   & $\cellcolor{secondcol}{67.3}$ & $80.8$ &  $74.6$ & \quad & - & - & - & - & \quad & - & - & - & - \\
      UMPD \cite{umpd} & $\cellcolor{firstcol}{76.6}$ & $61.2$ & $90.1$ & $\cellcolor{firstcol}{86.0}$  & \quad & $\cellcolor{secondcol}{73.8}$ & $\cellcolor{secondcol}{59.0}$ & $\cellcolor{secondcol}{88.6}$ & $\cellcolor{secondcol}{84.8}$ & \quad & $\cellcolor{secondcol}{67.5}$ & $\cellcolor{secondcol}{79.4}$ & $\cellcolor{firstcol}{93.4}$ & $\cellcolor{secondcol}{72.6}$ \\
      \midrule[0.4pt]
      \bf{DCHM} & $\cellcolor{firstcol}{84.2}$ & $\cellcolor{firstcol}{80.3}$ & $\cellcolor{secondcol}{90.2}$ & $\cellcolor{secondcol}{84.6}$  & \quad & $\cellcolor{firstcol}{80.1}$ & $\cellcolor{firstcol}{73.9}$ & $\cellcolor{firstcol}{91.2}$ & $\cellcolor{firstcol}{88.7}$ & \quad & $\cellcolor{firstcol}{78.4}$ & $\cellcolor{firstcol}{82.3}$ & $\cellcolor{secondcol}{90.7}$ & $\cellcolor{firstcol}{86.9}$ \\
      \bottomrule
      \end{tabular}
      }
  \end{center}
\vspace{-12pt}
  \caption{Performance (\%) comparison of label-free methods on Wildtrack, Terrace and MultiviewX dataset.}
  \label{table:unsupervised_comparison}
  \vspace{-10pt}
\end{table*}


\subsection{Evaluation}

\noindent  \textbf{Comparison of depth estimation methods for human modeling}. 
We evaluate several monocular depth estimation methods including Marigold~\cite{Marigold}, Metric3D v2~\cite{metric3d}, Depth Anything v2~\cite{depth_anything_v2}, and Depth Pro~\cite{depthpro}, as well as the framework Mast3R~\cite{mast3r} as the baseline for human modeling. 
Monocular depth reconstruction requires scaling relative depth $d_{rel}$ for accurate fusion. Specifically, we compute the metric depth $d_{metric}$ using camera calibration and known ground distance. The scale factor is given by $scale = median(\frac{M_g d_{metric}}{M_g d_{rel}})$, where $M_g$ is ground mask and $median(.)$ denotes the median operation. The final scaled depth is $d_{rel} \times scale$. . Point clouds are then fused into a global coordinate system via projection. Unlike monocular depth methods, Mast3R leverages image pairs to predict point maps and estimate camera poses via point cloud registration. We align these poses with dataset camera poses to refine scale and range.
As shown in Figure.~\ref{fig:cover}, our method achieves accurate multiview fusion even in occluded scenes. 
In contrast, existing methods struggle with consistent fusion leading to misalignment (e.g., the two seated individuals in Figure.~\ref{fig:cover}), geometry errors and increased false positives in localization. We predict point clouds from using methods and initialize them to obtain Gaussians for localization. Reconstruction quality is evaluated indirectly via detection metrics. Table~\ref{table:depth_comparison} demonstrates that our method accurately fuses depth across views.

\noindent \textbf{Comparison with label-free methods}.
We compare with label-free methods~\cite{xu2016multi, lopez2022semantic, zhu2019multi, Lima_2021_CVPR, umpd} in Table~\ref{table:unsupervised_comparison}.
The results demonstrate that DCHM achieves state-of-the-art performance in the Terrace dataset, significantly outperforming all other methods. It remains highly competitive on Wildtrack and MultiviewX, demonstrating superior performance in Wildtrack with MODP ($80.3\%$), exceeding UMPD by $31.2\%$, and in MultiviewX with MODA ($78.4\%$), surpassing UMPD by $16.1\%$. This precise localization is primarily attributed to accurate human modeling, highlighting the effectiveness of our framework.

\noindent \textbf{Deep comparison with concurrent work UMPD~\cite{umpd}}. In human modeling, UMPD and our method share a similar framework, performing 2D segmentation followed by 3D reconstruction. Our main contribution lies in 3D reconstruction, leveraging accurate human modeling. We decompose UMPD’s ``\textit{SIS}'' and ``\textit{GVD+VBR}'' for fair comparison. With the same segmentation, our reconstruction consistently outperforms UMPD’s by mitigating sparse view inconsistency, reduce erroneous reconstruction (see Figure~\ref{fig:cover}).



\noindent \textbf{Comparison with Label-Based Methods}. Supersingly, integrating our human modeling with ~\cite{mvchm} achieves SOTA performance in label-based methods with 3D annotation. See the supplementary for details.

\noindent \textbf{Inference Speed.} In label-free settings, our method achieves a real-time inference speed of 1.2 FPS while surpassing UMPD~\cite{umpd} in detection accuracy. See the supplementary for additional details.

\subsection{Ablation study}
\begin{figure}[t]
  \centering{
\includegraphics[width=0.9\linewidth]{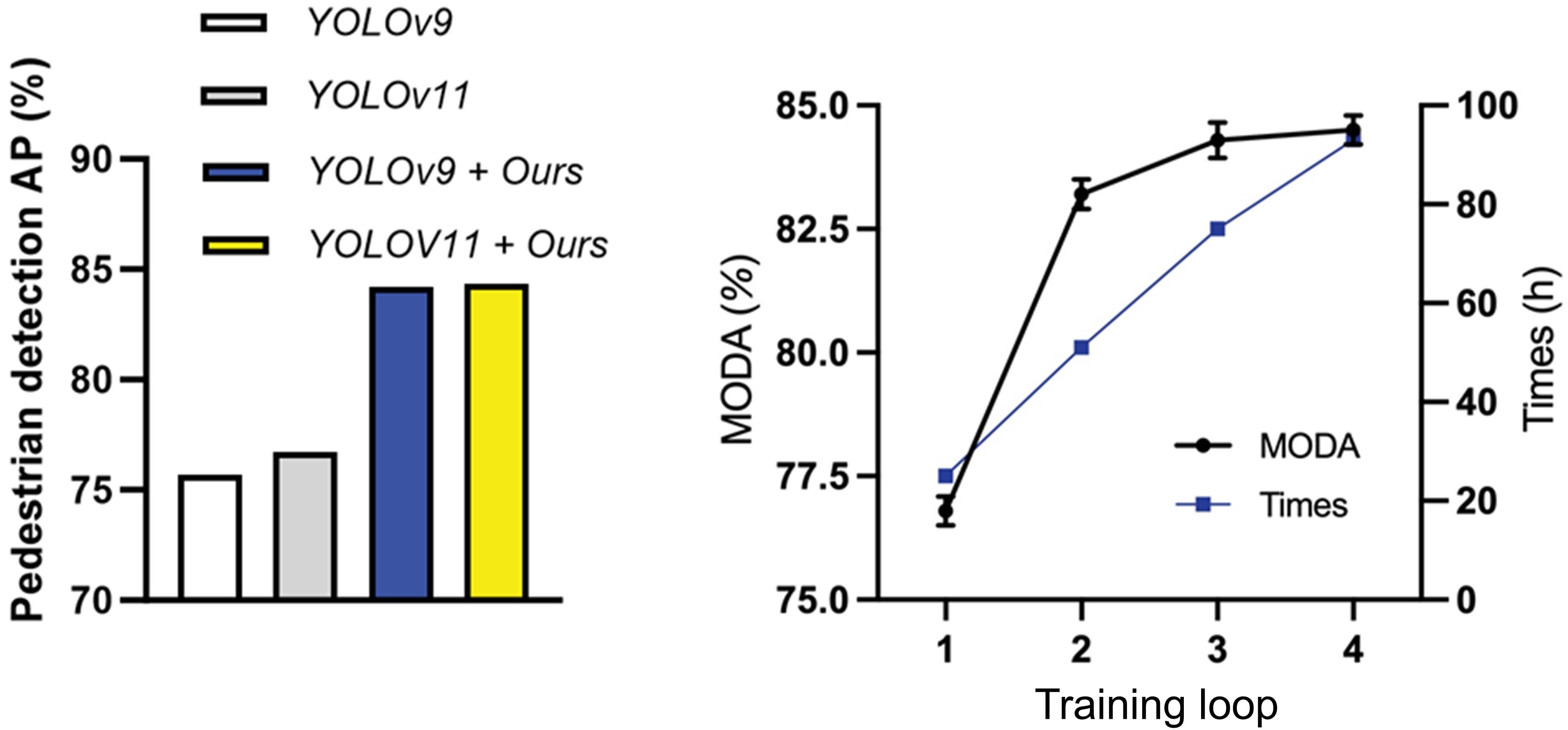}}
\vspace{-5pt}
  \caption{\textbf{Ablation study}. 
  \textit{Left}: Pedestrian detection AP (\%) for YOLOv9, YOLOv11, and their enhanced versions (“\textit{Ours}”), showing improved performance. \textit{Right}: MODA (\%) and training time (hours) across loops, illustrating the accuracy-cost trade-off.}   
  \label{fig:ablation_study}
\vspace{-10pt}
\end{figure}

\noindent  \textbf{Importance of detection compensation}.
Off-the-shelf detectors often miss occluded detections. Existing multiview methods~\cite{mvchm, mvdet, shot, mvdetr} use fusion to improve results, yet missed detections in single views persist. We are the \textit{first} to directly address this by introducing multiview compensation, recovering previously missed targets in final detections (see Figure~\ref{fig:label_compensation}.D). To fairly assess the effectiveness of our approach, we compare YOLOv9~\cite{yolov9} and YOLOv11~\cite{yolov11} with and without detection compensation on the Wildtrack+ dataset~\cite{mvchm}. As shown in the Figure~\ref{fig:ablation_study}.~\textit{left}, incorporating detection compensation enhances detection performance.

\noindent \textbf{Impact of training loop}. 
The training loop includes pseudo-depth generation, mono-depth fine-tuning, and multiview detection compensation. As shown in Figure~\ref{fig:pseudo_label}, the iterative framework progressively increases valid pseudo-depth, improving detection accuracy (Figure~\ref{fig:ablation_study}.~\textit{right}). However, the training time is also extended, as GS needs to optimize to obtain the pseudo-depth label for each frame.  By the third training loop, accuracy improvements become minimal; therefore, we limit training to three loops to balance accuracy and cost.

\noindent \textbf{The impact of optimization input}. As discussed above, pixel-wise supervision causes overlapping region inconsistency in sparse views. Superpixels aggregate local features to mitigate this issue. Table~\ref{tab:pixel_vs_superpixel} shows that superpixel-wise supervision enhances reconstruction, improving detection.

\begin{figure}[t]
    \centering
    \begin{minipage}{0.55\linewidth}
        \centering
        \resizebox{\linewidth}{!}{
            \begin{tabular}{ccc}
                \toprule
                2D Segmentation & 3D Reconstruction & MODA \\ 
                \midrule[0.2pt]
                \textit{SIS}~\cite{umpd} & \textit{GVD+VBR}~\cite{umpd} & $76.6$ \\ 
                YOLOv11~\cite{yolov11} & \textit{GVD+VBR}~\cite{umpd} & $78.9$ \\ 
                \textit{SIS}~\cite{umpd} & \textit{Ours} & $82.5$ \\ 
                YOLOv11~\cite{yolov11} & \textit{Ours} & $\mathbf{84.2}$ \\ 
                \bottomrule
            \end{tabular}
            }
        \captionof{table}{Deep comparison with UMPD~\cite{umpd} in 2D and 3D.}
        \label{tab:comparison}
    \end{minipage}
    \hfill
    \begin{minipage}{0.4\linewidth}
        \centering
        \resizebox{\linewidth}{!}{
            \begin{tabular}{cc}
                \toprule
                Optimization input &  MODA \\ 
                \midrule[0.2pt]
                Pixel-wise & $71.7$ \\ 
                Superpixel-wise & $\mathbf{84.2}$ \\ 
                \bottomrule
            \end{tabular}   
        }
        \captionof{table}{Comparison with optimization input.}
        \label{tab:pixel_vs_superpixel}
    \end{minipage}
\vspace{-15pt}
\end{figure}
\vspace{-1em}
\section{Conclusion}
We presented Depth-Consistent Human Modeling (DCHM), a new framework for robust multiview human detection. By leveraging superpixel-based Gaussian Splatting and geometric constraints for mono-depth iterative refinement, DCHM achieves accurate human modeling in sparse views, occluded scene. The accurate reconstruction can map per-view independent pedestrian cross views and achieve consistent segmented Gaussian in 3D. Extensive evaluations demonstrate that our DCHM approach outperforms existing label-free methods, and becomes a promising solution for accurate, consistent multiview pedestrian detection.

{
    \small
    \bibliographystyle{ieeenat_fullname}
    \bibliography{main}
}

\clearpage
\setcounter{page}{1}
\maketitlesupplementary

\begin{algorithm}[H]
\caption{Multi-view Label Matching}
\label{sup:matching}
\begin{algorithmic}[1]
\small
\Statex $G$: Set of optimized Gaussians from final training cycle
\Statex $M^v$: Set of pedestrian masks for each camera view $v$
\Statex $V$: View configuration of the current camera
\Statex $\tau_{vis}$: Visibility threshold
\Statex \hrulefill
\Function{LabelMatching}{$G, M^v, V, \tau_{vis}$}
    \State $G_{mask} \gets \text{RasterizeGaussianID}(G, V)$
    \For{$M_i^v$ in $M^v$}
        \If{$\exists g \in (G_{mask} \cap M_i^v) : g \text{ has ID}$}
            \State $c\_id \gets \text{MostFrequentID}(G_{mask} \cap M_i^v)$
            \State $\text{AssignID}(G_{mask} \cap M_i^v, M_i^v, c\_id)$
        \Else
            \State $n\_id \gets$ GenerateNewID()
            \For{\textbf{all} $g$ in $(G_{mask} \cap M_i^v)$}
                \State $w \gets \text{ComputeBlendingWeight}(g)$
                \If{$w > \tau_{vis}$}
                    \State $\text{AssignID}(G_{mask} \cap M_i^v, M_i^v, n\_id)$
                \EndIf
            \EndFor
        \EndIf
    \EndFor
    \State \Return $G, M^v$
\EndFunction
\end{algorithmic}
\end{algorithm}

\begin{algorithm}[H]
\caption{Multi-view Clustering and Localization}
\label{sup:clustering}
\begin{algorithmic}[1]
\small
\Statex $\tilde{G}$: Set of Gaussians with assigned IDs
\Statex $\tau_{cluster}$: Threshold for minimum number of Gaussians
\Statex $\epsilon$: DBSCAN epsilon parameter
\Statex $minPts$: DBSCAN minimum points parameter
\Statex \hrulefill

\Function{Cluster}{$\tilde{G}, \tau_{cluster}, \epsilon, minPts$}
    \State $L \gets \{\}$ \Comment{Initialize empty set of locations}
    \For{each unique ID $i$ in $\tilde{G}$}
        \State $\tilde{G}_i \gets \{g \in \tilde{G} : g \text{ has ID } i\}$
        \If{$|\tilde{G}_i| > \tau_{cluster}$}
            \State $C_i \gets \text{DBSCAN}(\tilde{G}_i, \epsilon, minPts)$ \Comment{Apply DBSCAN clustering}
            \For{each cluster $c$ in $C_i$}
                \State $cnt_c \gets \text{ComputeCenter}(c)$
                \State $conf_c \gets |c|$ \Comment{Confidence is the number of Gaussians}
                \State $L \gets L \cup \{(cnt_c, conf_c)\}$
            \EndFor
        \EndIf
    \EndFor
    \State $L \gets \text{NMS}(L)$ \Comment{Apply Non-Maximum Suppression}
    \State \Return $L$ \Comment{Return final predicted locations}
\EndFunction
\end{algorithmic}
\end{algorithm}

\begin{algorithm}[t]
\caption{Multi-view Unsupervised Localization}
\label{sup:unsup_loc}
\begin{algorithmic}[1]
\small
\Statex $G$: Set of optimized Gaussians from final training cycle
\Statex $M^v$: Set of pedestrian masks for all camera views $v$
\Statex $V$: Set of view configurations for all cameras
\Statex $\tau_{vis}$: Visibility threshold
\Statex $\tau_{cluster}$: Minimum Gaussian count for each cluster
\Statex $\epsilon$: DBSCAN epsilon parameter
\Statex $minPts$: DBSCAN minimum points parameter threshold
\Statex \hrulefill
\For{\textbf{all} camera $V^c$ in $V$} \Comment{Label matching}
    \State $\tilde{G}, M^v \gets \text{LabelMatching}(G, M^v, V^c, \tau_{vis})$
\EndFor
\State $L \gets \text{Cluster}(\tilde{G}, \tau_{cluster}, \epsilon, minPts)$ \Comment{Clustering and localization}
\end{algorithmic}
\end{algorithm}

\begin{algorithm}[t]
\caption{Detection Compensation}
\label{sup:detect_compensation}
\begin{algorithmic}[1]
\small
\Statex $r$: Reference view (index) $r$
\Statex $D$: Predicted depth of camera view 
\Statex $K$: Intrinsic matrix of camera view
\Statex $Rt$: Extrinsic matrix of camera view
\Statex $I$: Image of camera view
\Statex $M^v$: Foreground mask of camera view
\Statex $\tau_{pcs}$: Threshold for the number of valid projection points
\Statex $\tau_{m}$: Threshold for the proportion of projection points outside the source view foreground mask
\Statex $\epsilon$: DBSCAN epsilon parameter
\Statex $minPts$: DBSCAN minimum points parameter threshold
\Statex \hrulefill
\Function{DetComp}{$r, D, K, Rt, I, M^v, \tau_{pcs}, \tau_{m}, \epsilon, minPts$}
    \For{source view $s$ in all cameras}
        \State ${u}^{s \shortto r} \gets \texttt{Project}(D^s, K^s, Rt^s, M^{sv}, K^r, Rt^r)$ 
        \Comment{Source view projections on reference view $r$}
        \State $\tilde{{u}}^{s \shortto r} \gets \texttt{DBSCAN}({u}^{s \shortto r}, \epsilon, minPts)$ 
        \Comment{Filter outliers}
        
        \If{$|\tilde{{u}}^{s \shortto r}| < \tau_{pcs}$} \texttt{Continue} \EndIf 
        \Comment{Check projection count}
        
        \If{$\frac{|\tilde{{u}}^{s \shortto r}\cap M^r|}{|M^r|} > \tau_{m}$} \texttt{Continue} \EndIf
        \Comment{Skip on high overlap in new regions}

        \State $bbox \gets \texttt{BBox}(\tilde{{u}}^{s \shortto r})$ 
        \Comment{Get projection bounds}

        \State $M^{r\prime} \gets \texttt{SAM}(\tilde{{u}}^{s \shortto r}, bbox)$
        \Comment{Initial mask}
        
        \State $\tilde{{u}}^{s \shortto r}, bbox \gets \texttt{Sample}(M^{r\prime})$
        \Comment{Refine prompts}
        
        \State $M^{r\prime}\gets \texttt{SAM}(\tilde{{u}}^{s \shortto r}, bbox)$
        \Comment{Final mask}
        
    \EndFor
\EndFunction
\end{algorithmic}
\end{algorithm}

\begin{table}[t]
\begin{center}
    \scriptsize
    \resizebox{0.85\linewidth}{!}{
    \begin{tabular}{c c c c}
    \toprule
    \textbf{Method} & \textbf{Label-based} & \textbf{Accuracy} (\textit{MODA}) & \textbf{Speed} ($FPS$) \\
    \midrule[0.4pt]
    UMPD~\cite{umpd} & \ding{55} & $76.6$ & $1.0$\\
    \textit{Ours}   &  \ding{55} & $84.2$ & $1.2$ \\
     \midrule[0.4pt]
    MVDet~\cite{mvdet} & \checkmark & $88.7$ & $5.3$\\
    3DROM~\cite{3DROM} & \checkmark & $93.9$ & $2.4$\\
    MvCHM~\cite{mvchm} & \checkmark & $95.3$ & $2.5$\\
    \textit{Ours} & \checkmark & $\textbf{95.5}$ & $\textbf{6.1}$ \\
     
    \bottomrule
    \end{tabular}
    }
\end{center}
\caption{ Comparison of accuracy and computational efficiency.}
\label{table:computational_cost_appendix}
\end{table}


\begin{table*}[t!]
  \begin{center}
      \scriptsize
      \resizebox{0.65\linewidth}{!}{
      \begin{tabular}{l cccc c cccc}
      \toprule
      \multirow{2}{*}{Method} & \multicolumn{4}{c}{\bf Wildtrack} & \quad & \multicolumn{4}{c}{\bf MultiviewX}\\
      \cmidrule[0.3pt]{2-5} \cmidrule[0.3pt]{7-10}
      & MODA & MODP & Precision & Recall & \quad & MODA & MODP & Precision & Recall\\
      \midrule[0.4pt]
      MVDet \cite{mvdet}   &  $88.7$   &  $73.6$   &  $93.2 $  & $95.4$ & \quad &  $83.9$  &  $79.6$  &  $96.8$  &  $86.7$  \\
      
      SHOT \cite{shot}  &  $90.8$   & $77.7$ & {$96.0$} &  $94.3$   & \quad &  $88.3$  &  $82.0$  &  $96.6$  &  $91.5$  \\
      MVDeTr \cite{mvdetr}  & $92.1$ & $84.1$ & $96.1$ &  $94.5$  & \quad & $93.7$ & $\cellcolor{secondcol}{91.3}$ & $\cellcolor{firstcol}{99.5}$ & $94.2$  \\
      3DROM \cite{3DROM} & $93.9$  &  $76.0$   & $\cellcolor{secondcol}{97.7}$& $96.2$ &\quad & $\cellcolor{secondcol}{95.0}$ & $84.9$ & $99.0$ & $\cellcolor{secondcol}{96.1}$\\
      MvCHM \cite{mvchm} & $\cellcolor{secondcol}{95.3}$  &  $\cellcolor{secondcol}{84.5}$   & $\cellcolor{firstcol}{98.2}$& $\cellcolor{secondcol}{97.1}$ &\quad & $93.9$ & $88.3$ & $98.5$ & $94.8$\\
      \midrule[0.45pt]
      \bf{Our-Sup} & $\cellcolor{firstcol}{95.5}$ & $\cellcolor{firstcol}{90.2}$ & $\cellcolor{firstcol}{98.2}$ & $\cellcolor{firstcol}{97.4}$ & \quad & $\cellcolor{firstcol}{95.1}$ & $\cellcolor{firstcol}{91.5}$ & $\cellcolor{secondcol}{99.1}$ & $\cellcolor{firstcol}{96.8}$ \\
      \bottomrule
      \end{tabular}
      }
  \end{center}
  \vspace{-12pt}
  \caption{Performance (\%) of \textbf{\textit{supervised}} methods on Wildtrack and MultiviewX. \textbf{Our-Sup}: our human modeling + \textit{supervised} localization.}
  \label{table:supervised_comparison}
\end{table*}

\section{Supervised localization} 
\label{sec:sup_loc}
We conducted additional experiments to explore if our depth-consistent human modeling can enhance label-based methods.

\noindent\textbf{Methodology}. For supervised localization, we adopt the approach from MvCHM~\cite{mvchm}, where pedestrians are represented as point clouds and an off-the-shelf detection framework is employed. We refer to this process as \textit{label-based aggregation}. To aggregate features from multiple views, we convert point clouds into feature vectors using the network from \cite{pointpillars}, concatenating them to regress pedestrian positions on the ground plane. 
We discretize the point clouds into a grid on the BEV plane, creating pillars (vertical voxels \cite{pointpillars}). We then use PointNet \cite{pointnet} to extract high-dimensional pillar features. Following the methodology from \cite{voxelnet}, these features are flattened into the BEV plane for the final position regression. Pedestrian occupancy is represented as Gaussian maps, and we employ focal loss \cite{focal_loss} for position regression, defined as:
\vspace{-5pt}
\begin{equation}
    \mathcal{L}_{reg} = -\alpha (1 - p)^\gamma \log(p),
    \label{eq:focal_loss}
\vspace{-5pt}
\end{equation}
where $\alpha$ and $\gamma$ are hyperparameters specified in \cite{focal_loss}.

\noindent
\textbf{Comparison}. We replace label-free clustering with label-based aggregation to explore the potential of our framework under supervised settings. Quantitative results on the Wildtrack and MultiviewX datasets are reported in Table~\ref{table:supervised_comparison}. Our method (``\textbf{Ours}'') outperforms others across key metrics. Specifically, our method achieves the highest MODA ($95.5\%$) and MODP ($90.2\%$), indicating superior detection accuracy and localization precision. While MvCHM~\cite{mvchm} also models pedestrians using point clouds, it relies on the accurate detection of human standing points. In contrast, our method does not require keypoints for localization. Instead, we achieve a dense representation of pedestrians through consistent monocular depth estimation, which significantly aids downstream tasks such as multiview detection.

\section{Unsupervised localization}
Our multiview unsupervised localization is summarized in Algorithm~\ref{sup:unsup_loc}. The main functions, such as \textit{matching} and \textit{clustering}, are listed in Algorithm~\ref{sup:matching} and~\ref{sup:clustering}.

\section{Multi-view Detection Compensation}
We formalize our multiview detection compensation approach in Algorithm~\ref{sup:detect_compensation}.

\section{Computational cost} 
We provide an additional comparison of accuracy (\textit{MODA}) and computational efficiency in Table~\ref{table:computational_cost_appendix}. 
Integrating proposed human modeling with supervised method outperforms existing label-based approaches, achieving both the highest accuracy and computational speed.

\section{Ground depth calculation}
This section explains the process of calculating ground depth in the Wildtrack dataset. Although the ground range differs in the MultiviewX dataset, the calculation method remains consistent across both datasets. 
The ground plane is specified in a world coordinate system with an $x$-range $[0, 480]$, and a $y$-range $[0, 1440]$. $z$ is set to $0$ by default. We uniformly sample points $\mathbf{p}^w \in \mathbb{R}^3$ on the ground within the predefined range. 
Given the camera $c$ specified via the rotation matrix $\mathbf{R}^c$ and the translation vector $\mathbf{t}^c \in \mathbb{R}^3$ in the world coordinate system, the ground points $\mathbf{p}^w$ in the world coordinate system are transformed into the camera coordinate system as:
\begin{equation}
d^c = [\mathbf{p}^c]_z, \ \ \ \  \mathbf{p}^c = (\mathbf{R}^c)^{-1} ( \mathbf{p}^w - \mathbf{t}^c ),
\label{eq:ground_depth}
\end{equation}
where $[\,\cdot\,]_z$ denotes the $z$-coordinate of the point in the camera coordinate system. This process is repeated for each camera to calculate the corresponding ground depth.

\section{GS depth filtering details}
This section provides additional details about the GS depth filtering process. We define $\mathbf{u}^s$ as the homogenous pixel coordinate in source view $v^s$, $\mathbf{K}^{s}$ is the source view camera intrinsic matrix. We unproject each pixel $\mathbf{u}^s$ from source view $v^s$ into a 3D point $\mathbf{p}^{s \rightarrow w} \in \mathbb{R}^3$ in world coordinate, using predicted depth map $D^s(\mathbf{u}^s)$ and camera poses:
\begin{equation}
    \mathbf{p}^{s \shortto w} = \mathbf{R}^s (D^s(\mathbf{u}^s) (\mathbf{K}^{s})^{-1} \mathbf{u}^s) + \mathbf{t}^s.
\end{equation}
Here, $\mathbf{R}^s$ is the $3 \times 3$ camera rotation matrix, and $\mathbf{t}^s \in \mathbb{R}^3$ is the camera translation vector. With reference view camera pose $\mathbf{R}^r$ and $\mathbf{t}^r$, we can project $\mathbf{u}^s$ to reference view $v^r$ and obtain $D^{s \shortto r}$via:
\begin{equation}
    D^{s \shortto r}(\mathbf{u}^{s \shortto r}) = \left[\mathbf{K}^r (\mathbf{R}^{r})^{-1}(\mathbf{p}^{s \shortto w} - \mathbf{t}^r)\right]_z,
\end{equation}
\begin{equation}
    \mathbf{u}^{s \shortto r} = \mathbf{K}^r (\mathbf{R}^{r})^{-1}(\mathbf{p}^{s \shortto w} - \mathbf{t}^r) / D^{s \shortto r}(\mathbf{u}^{s \shortto r}).
\end{equation}

\section{Discussion}

\begin{figure}
  \centering{
    \includegraphics[width=0.8\linewidth]{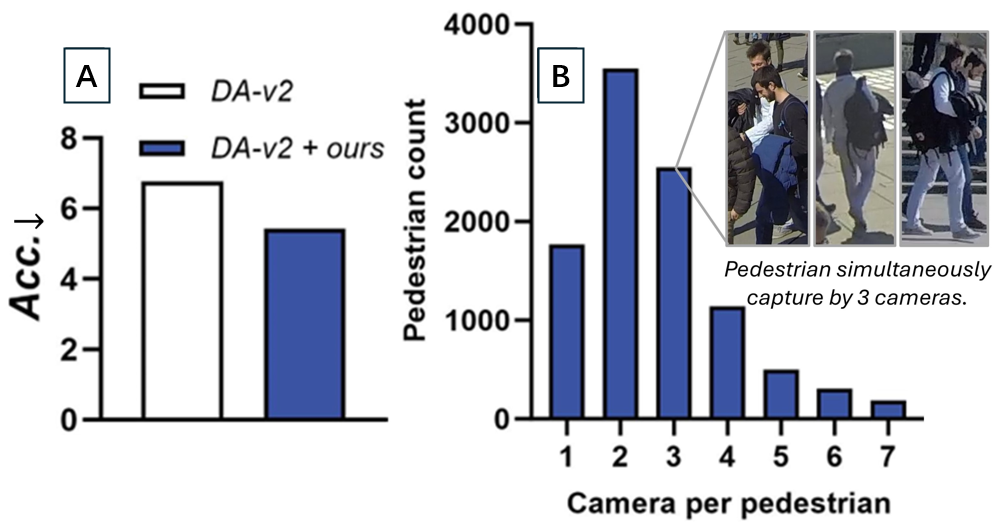}}
  \vspace{-5pt}
  \caption{(\textbf{A}) Error comparison with DA-v2 on DTU. (\textbf{B}) Histogram of the number of pedestrians captured simultaneously by multiple cameras.}
  \vspace{-5pt}
  \label{fig:sup_exp}
\end{figure}

\noindent \textbf{Sparse view setup in multi-view detection differs from existing sparse view reconstruction in several aspects.} 1.~\textit{Large-scale scene}: unlike traditional indoor or object‐level settings, we tackle crowded surveillance (e.g., the Wildtrack dataset) covering a $30\,\mathrm{m}\times40\,\mathrm{m}$  plaza with over $30$ pedestrians per frame. 2. \textit{Severe occlusion}: despite of seven cameras, severe occlusion causes most pedestrians ($78.67 \%$) to appear in only one to three views (see Fig.~\ref{fig:sup_exp}.B) resulting in an extremely challenging sparse-view setup.
Additionally, even though some pedestrians are observed in multiviews, their appearances are dramatically different across views (see Fig.~\ref{fig:sup_exp}.B), making the problem even harder. In such challenging scenarios, generating accurate, consistent depth annotations for fine-tuning is exceptionally difficult. Therefore, our proposed method addresses a meaningful and underexplored challenge.

\begin{figure}
  \centering{
    \includegraphics[width=0.95\linewidth]{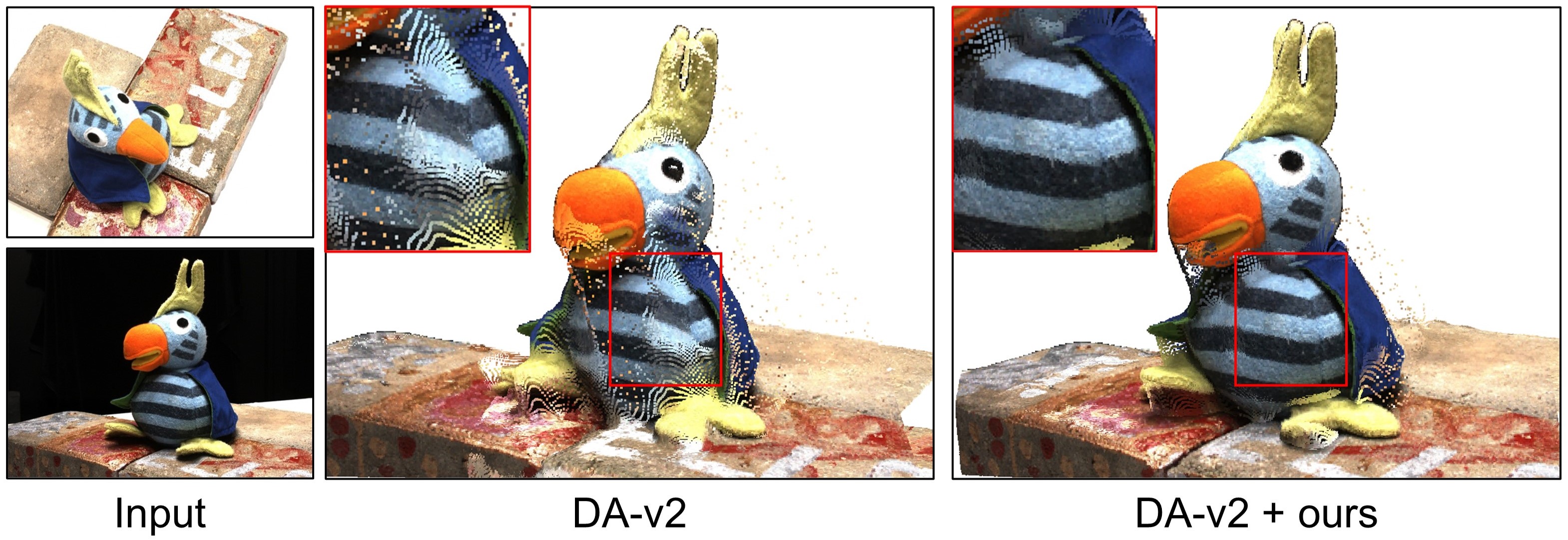}}
  \vspace{-5pt}
  \caption{Visual comparison of DepthAnything-v2 (DA-v2) with and without our optimization.}
  \vspace{-5pt}
  \label{fig:dtu}
\end{figure}

\noindent \textbf{DCHM also gains in multi-view depth estimation.} To quantify its impact on multi-view depth estimation, we evaluate on the DTU benchmark using DepthAnything-v2 (DA-v2)~\cite{depth_anything_v2} as the baseline. Under sparse-view inputs, DA-v2 predicts relative depth per view, scales it using ground-truth, and fuses results into point clouds via known intrinsics and extrinsics. We finetune DA-v2 using pseudo-depth labels from our framework. Our approach yields more coherent depth maps and smoother fusion as in Fig.~\ref{fig:dtu}; and  lower depth error as in Fig.~\ref{fig:sup_exp}.B compared to the baseline. While designed for~\emph{Multiview Detection}, our method still improves multi-view depth estimation.

\noindent \textbf{Comparison: Ours \textit{vs.} Gaussian-based 3D Segmentation.} 
Gaussian splatting provides a promising approach for 3D segmentation. Methods like Gaussian Grouping~\cite{Gaussian_grouping} use video tracking to enforce 2D mask consistency across views before projecting to 3D, while Gaga~\cite{gaga} leverages spatial information to associate object masks across multiple cameras. However, these methods require densely overlapping camera views, which limits their applicability in scenarios with sparse view coverage.
Our task introduces greater challenges due to minimal scene overlap, making geometric reconstruction particularly demanding. Approaches relying on Gaussian splatting for reconstruction followed by segmentation assignment often struggle under these conditions. In contrast, our framework overcomes these limitations, facilitating effective scene reconstruction and extending 3D segmentation from object-centric domains to large-scale outdoor environments.

\section{Limitations and future works}
\noindent \textbf{Impact of depth estimation on detection compensation.} 
While multiview detection compensation improves monocular detection, its performance is highly dependent on the accuracy of depth estimation. Error in depth estimation can result in imprecise prompts for SAM, leading to noisy segmentation despite mask-guided sampling. Future work could focus on methods for assessing the validity of compensatory masks to mitigate these issues. 

\noindent \textbf{Lack full use of temporal information.}
Our method, which solely relies on multiview geometry for monocular depth fine-tuning, can lead to instability when objects are visible to only a single camera view. Incorporating temporal information could provide a more robust and consistent approach to depth estimation in such scenarios.

\begin{figure*}[t]
    \centering
    \vspace{-0.5cm} 
    \includegraphics[width=0.7\textwidth]{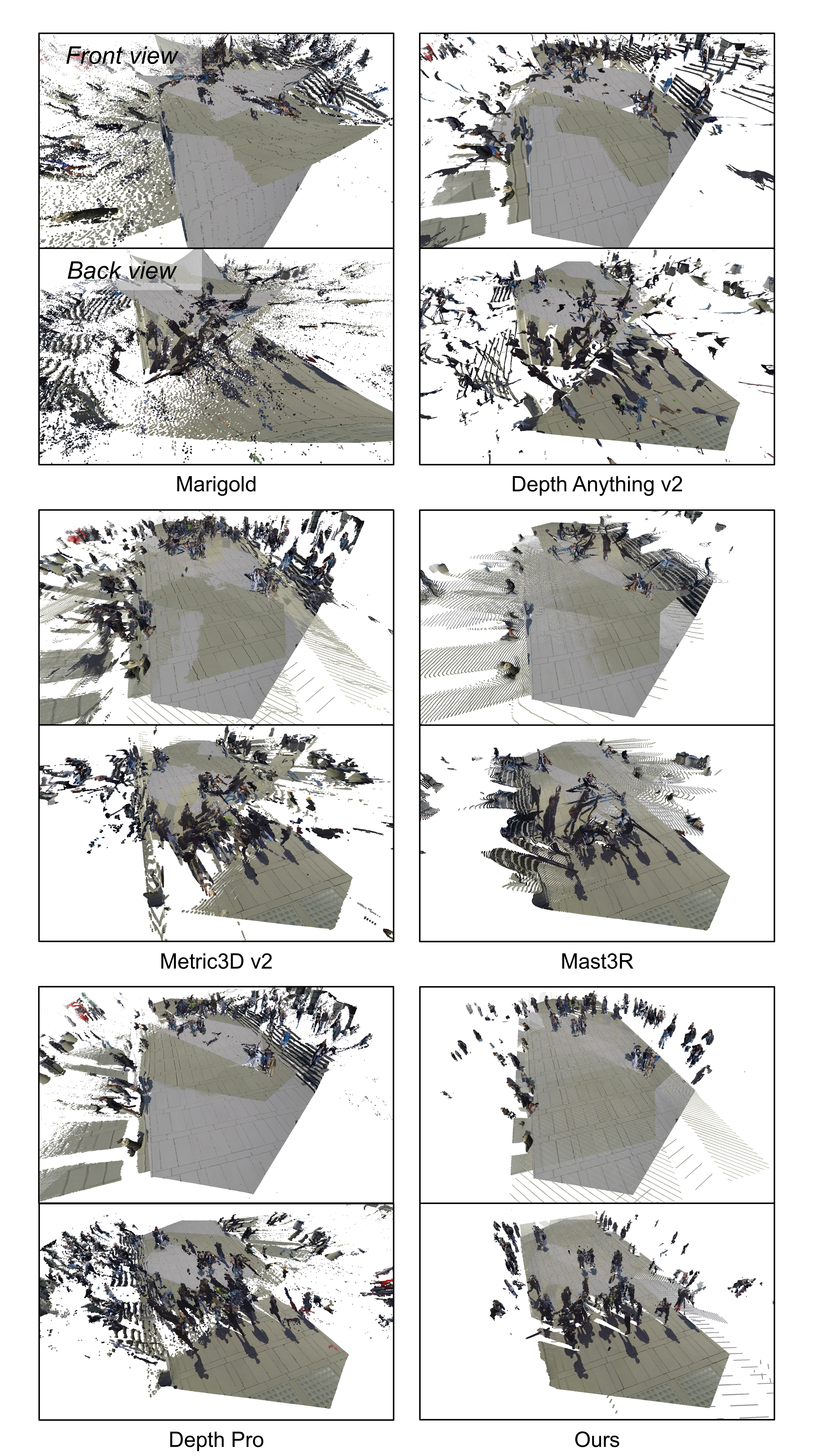}
    \vspace{-0.2cm}
    \caption{\textbf{Enlarged 3D reconstruction from various methods}. We present enlarged views of 3D reconstructions generated by baselines~\cite{Marigold, depth_anything_v2, metric3d, mast3r, depthpro}, showcasing both front and back perspectives for each. 
    Our method (``Ours'') produces more accurate and complete reconstructions compared to the baselines.}
    \label{fig:pointcloud}
\end{figure*}

\begin{figure*}[t]
    \centering
    \vspace{-0.5cm} 
    \includegraphics[width=0.99\textwidth]{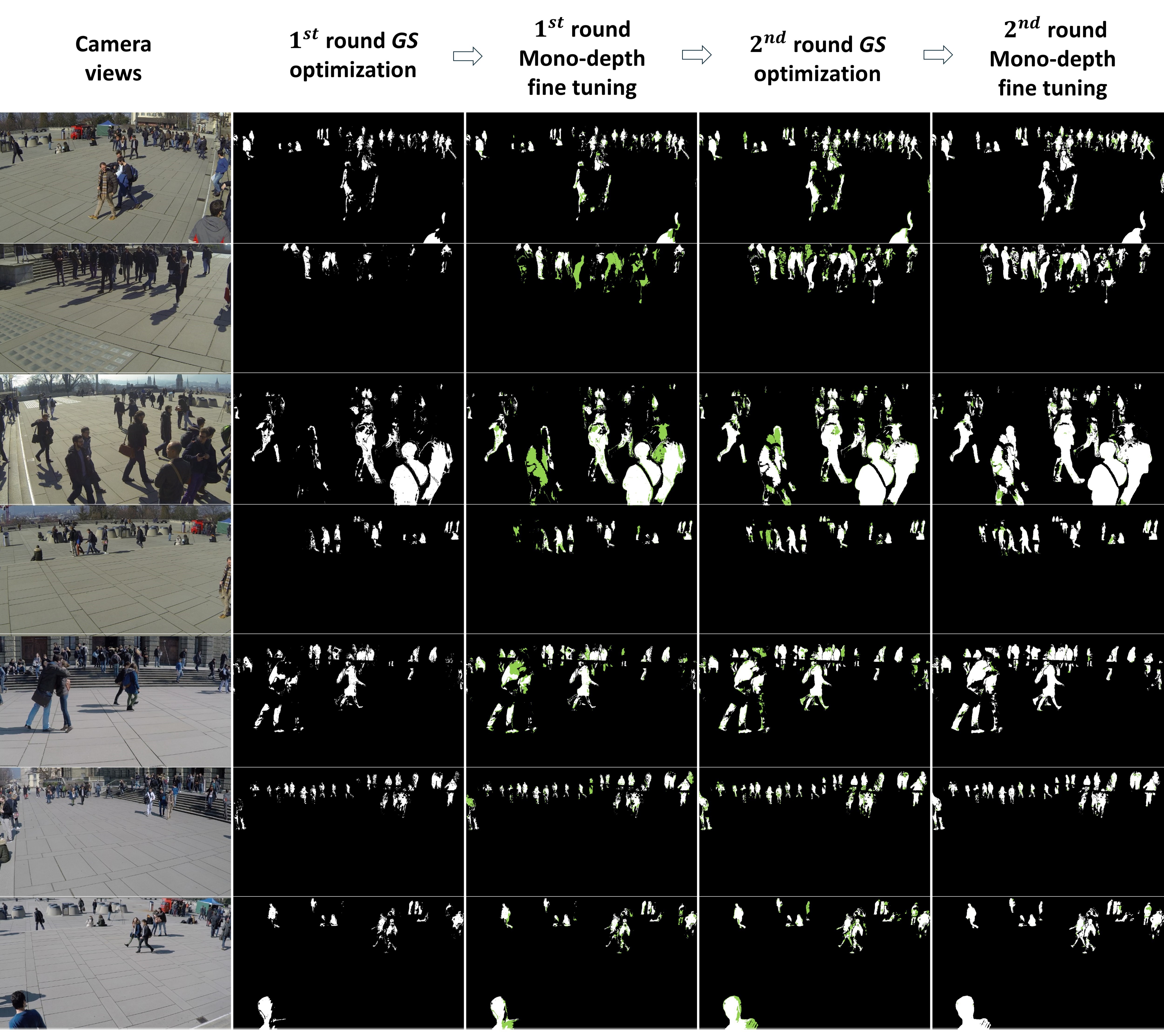}
    \vspace{-0.2cm}
    \caption{\textbf{Visualization of increasing valid pseudo-depth across all cameras}. We illustrate the progression of valid depth regions for all viewpoints during each round of optimization. This serves as a supplement visualization for Figure~\ref{fig:pseudo_label} in the main text.}
    \label{fig:pseudo_laebl_appendix}
\end{figure*}

\end{document}